%% file: main_cready.tex
\documentclass{article} %
\usepackage[dblblindworkshop,final]{neurips_2025}

\usepackage{times}

\usepackage{url}

\usepackage{CJKutf8}
\usepackage{arydshln}
\usepackage{booktabs}       %
\usepackage{amsfonts}       %
\usepackage{nicefrac}       %
\usepackage{microtype}      %
\usepackage{mathtools}
\usepackage{multirow}
\usepackage{bm}
\usepackage{epsfig}
\usepackage{graphicx}
\usepackage{subcaption}
\usepackage{amsthm}
\usepackage{amsmath}
\usepackage{amssymb}
\usepackage{enumitem}
\usepackage{makecell}
\usepackage{wrapfig}
\usepackage{indentfirst}
\usepackage{verbatim}
\usepackage{color}
\usepackage{setspace}
\usepackage{array}
\usepackage{stackengine}

\usepackage{algorithm}
\usepackage[algo2e,algoruled,boxed,lined]{algorithm2e}
\usepackage{algorithmicx}
\usepackage{algcompatible}

\usepackage{graphicx}
\usepackage[table]{xcolor}
\usepackage{anyfontsize}

\usepackage{caption}

\usepackage[pagebackref=true,breaklinks=true,colorlinks=true,bookmarks=false]{hyperref}
\definecolor{deepred}{HTML}{940000}
\hypersetup{linkcolor=deepred}
\hypersetup{urlcolor  = [rgb]{0.4,0.15,0.95}}
\hypersetup{citecolor=[rgb]{0.4,0.15,0.95}}

\definecolor{model}{HTML}{00639E}
\definecolor{opt}{HTML}{2A5F1C}

\usepackage{colortbl}
\definecolor{Gray}{gray}{0.94}
\newcolumntype{a}{>{\columncolor{Gray}}c}
\newcolumntype{Y}{>{\centering\arraybackslash}X}

\newlength\savewidth

\usepackage{tabularx}

\usepackage{pifont}
\usepackage{tocloft}
\usepackage[toc,page,header]{appendix}
\usepackage{adjustbox}
\usepackage{minitoc}

\usepackage[capitalize,noabbrev]{cleveref}

\usepackage{lipsum}

\usepackage{epigraph}

\usepackage{listings}

\lstdefinelanguage{yaml}{
  keywords={true,false,null,y,n},
  sensitive=false,
  comment=[l]{\#},
  morestring=[b]",
  morestring=[b]'
}

\workshoptitle{The Second Workshop on GenAI for Health: Potential, Trust, and Policy Compliance}

\title{Demo: Generative AI helps Radiotherapy Planning with User Preference}
\author{
  Riqiang Gao\textsuperscript{1}  \quad Simon Arberet\textsuperscript{1} \quad Martin Kraus\textsuperscript{1} \quad Han Liu\textsuperscript{1} \quad Wilko F.A.R. Verbakel\textsuperscript{2} \\ \textbf{Dorin Comaniciu}\textsuperscript{1} \quad \textbf{Florin C. Ghesu}\textsuperscript{1} \quad \textbf{Ali Kamen}\textsuperscript{1}\\[1.5mm]
  \textsuperscript{1}Digital Technology and Innovation, Siemens Healthineers\\
  \textsuperscript{2}Varian Medical Systems, Siemens Healthineers\\
}

\begin{document}
\maketitle
\doparttoc 
\faketableofcontents
\vspace{-0.1in}
\begin{abstract}

Radiotherapy planning is a highly complex process that often varies significantly across institutions and individual planners. Most existing deep learning approaches for 3D dose prediction rely on reference plans as ground truth during training, which can inadvertently bias models toward specific planning styles or institutional preferences. 
In this study, we introduce a novel generative model that predicts 3D dose distributions based solely on user-defined preference “flavors”. These customizable preferences enable planners to prioritize specific trade-offs between organs-at-risk (OARs) and planning target volumes (PTVs), offering greater flexibility and personalization. Designed for seamless integration with clinical treatment planning systems, our approach assists users in generating high-quality plans efficiently. Comparative evaluations demonstrate that our method can surpasses the Varian RapidPlan$^\text{TM}$ model in both adaptability and plan quality in some scenarios. Demo video: \url{https://huggingface.co/Jungle15/DoseProposerDemo}. 

\end{abstract}

\section{Introduction}
\label{sec:intro}
Radiotherapy (RT) plays a pivotal role in cancer treatment, being applicable for about half of all cancer cases \cite{Huynh2020ArtificialOncology}. In recent years, deep learning has been integrated into numerous stages of the RT planning workflow, including tasks such as dose prediction \cite{kui2024review,Babier2020OpenKBP:Challenge,gao2023flexible,zhang2024dosediff}, fluence map generation \cite{Wang2020FluenceTherapy,arberet2025beam}, end-to-end including multileaf collimator (MLC) leaf sequencing \cite{gao2024multi,hrinivich2024clinical}, and dose calculation \cite{xing2020feasibility}. Among these, 3D dose prediction has gained particular attention due to its potential to streamline the planning process. By leveraging inputs like CT images, delineated structures such as planning target volumes (PTVs) and organs at risk (OARs), and configuration data, deep models can estimate full 3D dose distributions without relying on time-consuming optimization \cite{Babier2020OpenKBP:Challenge,gao2023flexible}. This capability has broad applications: it can inform objective functions during plan optimization \cite{Babier2022OpenKBP-Opt}, facilitate quality assurance \cite{Gronberg2023DeepPlans,Gronberg2023DeepCancers}, and serve as a foundation for fully automated, AI-driven planning systems \cite{fan2019automatic,gao2024multi}. 

RapidPlan$^\text{TM}$ \cite{varian2024rapidplan} of Varian is a knowledge-based treatment planning system that uses conventional machine learning and a library of prior high-quality plans to model the relationship between anatomy and dose distribution. It generates personalized dose-volume histograms (DVHs) \cite{Drzymala1991DOSE-VOLUMEHISTOGRAMS} and optimization objectives, improving plan consistency, quality, and efficiency across various treatment sites.

Despite its clinical popularity, RapidPlan has notable limitations: (1) it relies on DVH predictions, which can miss critical spatial dose details, reducing the accuracy of personalized optimization for OARs, especially in anatomically complex cases; (2) its PCA-based regression pipeline is typically trained on dozens of plans, limiting generalizability across institutions and patient populations. Users need to train institution-specific models to reflect guidelines and preferences.

Recent deep learning models enable direct 3D dose prediction from anatomy and contours, offering voxel-level accuracy and spatial awareness with large-scale data \cite{Babier2020OpenKBP:Challenge,gao2025automating}. This enhances personalization, interpretability, and supports adaptive, automated planning. However, user interaction with a single model remains unexplored. Moreover, dose prediction alone is not a deliverable plan, and while dose mimicking \cite{Babier2022OpenKBP-Opt} and end-to-end AI pipelines \cite{gao2024multi} have been studied in research settings, integration into commercial clinical platforms is still limited.

To address these gaps, we contribute the following in this work:

\begin{itemize}
\item We propose a two-stage training framework with a foundational dose decoder, improving the stability of training in complex scenarios.
\item To our knowledge, this is the first dose prediction model with interactive sliders, enabling real-time customization of trade-offs between target homogeneity and OAR sparing.
\item We integrate our AI model with a widely used clinical treatment planning system and demonstrate its effectiveness in generating high-quality radiotherapy plans.
\end{itemize}

\section{Method}

\subsection{Problem Definition and Intuition}
\label{sec:define}
The homogeneity index (HI) and conformity index (CI) are two main metrics to evaluate plan quality on PTVs. The CI evaluates how well the prescribed dose conforms to the shape and size of the PTV. The HI measures the uniformity of the dose distribution within the PTV. $\text{HI} = (D_{05} - D_{95}) /D_{50}$, and $\text{CI} = V_{\text{covered}}/V_{\text{PTV}}$. $D_{t}$ represents dose received by $t$\% of the PTV (i.e., (100-$t$)-th percentile). $V_{\text{PTV}}$ is volume size of the PTV, and $V_{\text{covered}}$ is volume size of the PTV receiving at least prescribed dose. On the other hand, OAR sparing (e.g., mean doses of OARs), is often inversely related to PTV coverage and uniformity. Traditional planning models often fail to capture diverse user preferences, requiring separate models for different styles.

Generative AI has demonstrated strong capabilities in producing images and videos based on user prompts \cite{rombach2022high,ramesh2022hierarchical,wan2025wan}. However, current AI-based dose prediction models largely overlook individual planning preferences. This work aims to reduce that gap. After dose prediction, we convert the 3D dose to optimization objectives inspired by \cite{gao2025automating} to generate deliverable plans (more in Appendix \ref{sec:dose2obj}). 


\subsection{Flexible Dose Proposer}
\begin{figure}
    \centering
    \includegraphics[width=\linewidth]{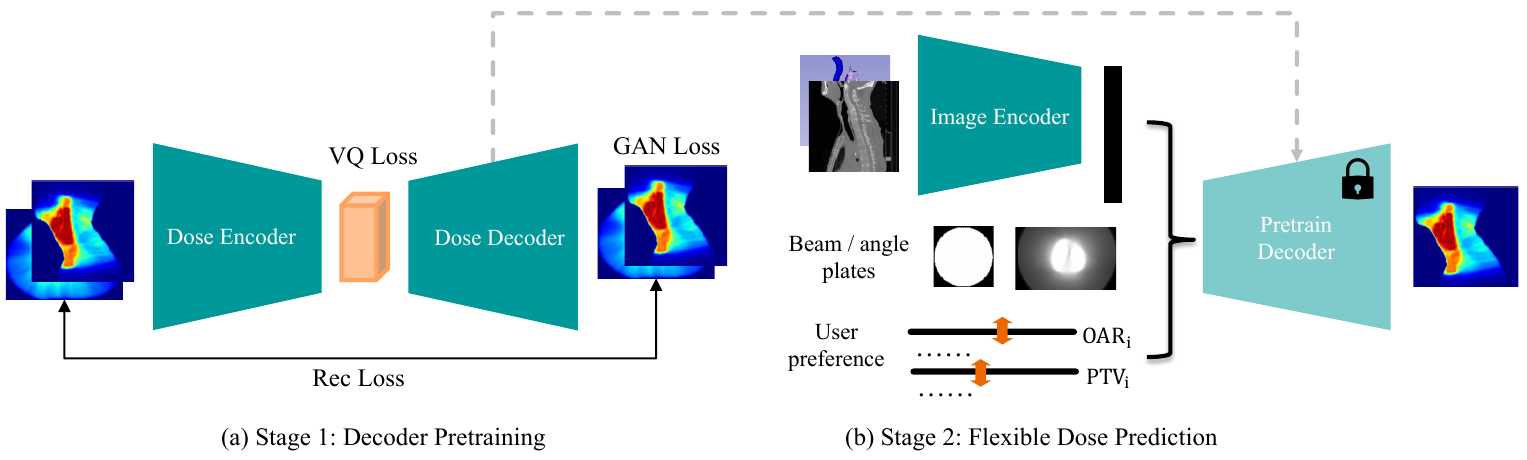}
    \caption{Our flexible dose proposer pipeline. Stage I pre-trains a foundational dose decoder, regularizing the dose prediction to be realistic. The second stage trains a flexible dose prediction model with heterogeneous conditions. }
    \vspace{-0.1in}
    \label{fig:framework}
\end{figure}
Our flexible dose proposer (FDP) is shown in Figure \ref{fig:framework}.  Stage I trains a foundational dose decoder using a VQ-VAE architecture \cite{van2017neural}. Stage II introduces multi-conditional inputs to enable context-aware dose prediction. For fast inference on 3D images, we deliberately exclude computationally intensive or iterative models such as diffusion-based approaches \cite{ho2020denoising}, although the core concept of FDP remains adaptable to other generative frameworks.   

\textbf{Stage I.} The VQ-VAE pretraining is inspired by the success of latent diffusion models \cite{rombach2022high,wan2025wan}, which use VAEs primarily for data compression to reduce inference time and computational cost. In contrast, our goal of Stage I pretraining is to stabilize Stage II training, especially under complex conditions like user preferences. Since we adopt one-step generation via GANs \cite{goodfellow2014generative}, our inference time is minimal (shown in the demo) compared to diffusion-based approaches. 
Denote real/reference data and prediction as $x$ and $\hat{x}$ respectively, the adversarial loss is defined as $L_{adv}(x, \hat{x}) = \mathbb{E}_{x}[(D(x) - 1)^2] + \mathbb{E}_{\hat{x}}[D(\hat{x})^2]$ and $L_{vq}$ is the vector quantisation loss following \cite{van2017neural}. 
To stabilize the training, we further introduce a uniform loss \cite{wang2020understanding} to regularize the latent space:  
\begin{equation}
\scriptsize 
\mathcal{L}_{\text{stage1}} = 
\underbrace{\mathbb{E}_i[\|x_i - \hat{x}_i\|]}_{\text{Reconstruction}} + 
\beta L_{vq} + L_{adv}(x, \hat{x})
 + 
\underbrace{\lambda \cdot \log \left( \mathbb{E}_{i < j} \left[ \exp(-t \|\hat{z}_i - \hat{z}_j\|^2) \right] \right)}_{\text{Uniformity}}.
\end{equation}
Model structure following MONAI \cite{cardoso2022monai} for VQVAE. Network parameters are in Appendix \ref{sec:structure}.

\textbf{Stage II.} We utilize the foundational dose decoder for Stage II to stabilize the training. The CT image and RT structures have been concatenated as multi-channel input for the image encoder \cite{gao2023flexible}. User preferences together with beam/angle plates, are encoded with adaptive instance normalization \cite{huang2017arbitrary}. The encoder structure follows the MedNext \cite{roy2023mednext} style, and network parameters are in Appendix \ref{sec:structure}. 

Loss functions for Stage II include reconstruction losses of image space $\{x, \hat{x}\}$ and latent space $\{z, \hat{z}\}$, adversarial loss of the image space, and objective loss $L_{\text{obj}}$, which are defined as follows: 
\begin{equation}
\footnotesize
\mathcal{L}^{(i)}_{\text{stage2}} = 
\underbrace{\|x_i - \hat{x}_i\|}_{\text{Image Reconstruction}} + 
\underbrace{\|z_i - \hat{z}_i\|}_{\text{Latent Reconstruction}} + 
L_{adv}(x_i, \hat{x}_i) + 
\mathcal{L}^{(i)}_{\text{obj}}.
\end{equation}
The $L_{\text{obj}}$ is the inconsistency penalty between the user preference prompts and the prediction metrics: 
\begin{equation}
\footnotesize
\mathcal{L}^{(i)}_{\text{obj}} = 
\underbrace{\|\Tilde{h} - \hat{h}\|}_{\text{ptv HI preference}} + 
\underbrace{\|p - \hat{p}\|}_{\text{ptv dose alignment}} + 
\underbrace{\|\Tilde{w} \cdot u_{\text{oar}} - \hat{u}_{\text{oar}}\|}_{\text{oar-sparing preference}},
\end{equation}
where $\Tilde{h}$ and $\hat{h}$ denote the homogeneity indices of user preference (i.e., slider value) and prediction, respectively; $p$ and $\hat{p}$ represent the PTV mean dose of the reference and predicted plans; $\Tilde{w}$ is the user-defined preference for OAR, and $u_{\text{oar}}$ is the OAR mean dose of the reference plan. For simplicity, we omit writing out $\mathcal{L}^{(i)}_{\text{obj}}$ for each structure.

During training, the sliding bar values $\{\Tilde{h}, \Tilde{w}\}$ are randomly sampled within predefined ranges. This sampling strategy enhances the alignment between user inputs and the model's predictions. In parallel, various loss regularizations—such as reconstruction loss and adversarial loss—are employed to guide learning. Additionally, Stage I pre-training introduces an extra layer of regularization. Together, these strategies encourage the model to integrate both patient-specific information (e.g., spatial distribution of PTV/OAR) and user interactions via the sliding bars into its dose predictions.
\vspace{-0.05in}
\section{Experiments} 
\begin{table}[]
    \centering
    \small
    \begin{tabular}{ccccccc}
    \toprule
       Cohort Index  & Cohort 0 &Cohort 1 &Cohort 2 &Cohort 3 &Cohort 4 &Cohort 5 \\
       \midrule
        Train/Valid/Test & 370/48/54 & 147/17/19 & 128/15/17& 103/14/12 & 52/8/7 & 20/1/4 \\
    \bottomrule
    \end{tabular}
    \caption{Data distribution used in Stage II training. The Stage I is pre-trained with 31K doses. }
    \vspace{-0.2in}
    \label{tab:data}
\end{table}
\textbf{Dataset}. We have included six cohorts of head-and-neck cancer cases, 
as shown in Table \ref{tab:data}. 
  The Stage I pre-training decoder has used a much larger scale dataset with 31K doses (not necessarily from high-quality plans), including head-and-neck, lung, prostate cancer cases. 

\begin{table}[b]
    \centering
    \tiny
    \begin{tabular}{c|@{\hspace{6pt}}c@{\hspace{6pt}}c@{\hspace{6pt}}c@{\hspace{6pt}}c@{\hspace{6pt}}c@{\hspace{6pt}}c@{\hspace{6pt}}c@{\hspace{6pt}}c}
    \toprule
        OAR & SpinalCord05 & Larynx-PTV & Lips & Mandible-PTV & OCavity-PTV & ParotidCon-PTV & ParotidIps-PTV & Esophagus \\
RP std & 3.02 & 3.49 & 2.54 & 2.93 & 4.04 & 3.38 & 2.96 & 3.21 \\ 
ours std & \textbf{1.18} & \textbf{1.78} & \textbf{1.09} & \textbf{1.27} & \textbf{1.82} & \textbf{1.22} & \textbf{1.78} & \textbf{1.27} \\
\rowcolor{gray!10}
RP mean & 3.25 & 6.29 & 4.71 & 6.41 & 7.37 & 4.51 & 5.54 & 2.43 \\
\rowcolor{gray!10}
ours mean & \textbf{1.69} & \textbf{2.94} & \textbf{2.06} & \textbf{2.01} & \textbf{4.13} & \textbf{1.92} & \textbf{2.92} & \textbf{1.18} \\
         \midrule
         OAR & SubmandL-PTV & Shoulders & SubmandR-PTV  & PosteriorNeck & PharConst-PTV & BrainStem 03 & Trachea & \textit{better count}\\
RP std  & 3.25 & 1.38 & 2.70 & 4.30 & 3.62 & 0.92 & 3.40 & \textit{0} \\
ours std  & \textbf{1.38} & \textbf{0.51} & \textbf{1.45} & \textbf{0.92} & \textbf{1.78} & \textbf{0.86} & \textbf{1.68} & \textit{\textbf{15}} \\
\rowcolor{gray!10}
RP mean  & \textbf{4.43} & 0.97 & 6.29 & 8.15 & 4.36 & 1.49 & 3.42 & \textit{1} \\
\rowcolor{gray!10}
ours mean  & 4.51 & \textbf{0.62} & \textbf{5.37} & \textbf{1.19} & \textbf{3.37} & \textbf{1.08} & \textbf{2.01} & \textit{\textbf{14}} \\
         \bottomrule
    \end{tabular}
    \caption{Statistics of intra-patient differences of expected and achieved DVHs ($ \downarrow $). RP: RapidPlan \cite{magliari2024hn}. ours: the predicted dose from FDP and objectives extracted from AI dose for Eclipse planning.}
    \label{tab:intra}
\end{table}
\textbf{Evaluation Metrics}. Dose-volume histograms (DVHs) \cite{Drzymala1991DOSE-VOLUMEHISTOGRAMS}  are graphical representations in RT planning, providing a quantitative assessment of dose distribution across target volumes and surrounding organs at risk. 
Given DVHs of a specific organ ($\text{DVH}_i$ has been normalized with same index of \texttt{Volume}, we compare the differences of \texttt{Dose}), intra-patient differences are defined as $D = \text{DVH}_{\text{expect}} - \text{DVH}_{\text{achieved}}$, where  $\text{DVH}_{\text{expect}}$ and $\text{DVH}_{\text{achieved}}$ are the expected and achieved DVHs for the organ. We define the metrics (mean and standard deviation) and and show illustartion about intra- and inter- patient differences at Appendix \ref{app:metrics}. 
Lower metric values represent the expected plan and achieved plan are closer, indicating better DVH estimations. 

We use a high-quality RapidPlan \cite{magliari2024hn,varian2024han} as baseline. The RT structure creation of all test plans followed the RapidPlan requirements. All the reference plans for AI are based on this RapidPlan.  

\textbf{Results.} Table \ref{tab:intra} and Table \ref{tab:inter} shows the statistics of intra-patient and inter-patient differences, respectively. In both scenarios, lower values especially lower standard deviation (std) indicate better estimations of DVHs. Our FDP model outperforms RapidPlan in both scenarios. 

\begin{figure}
\centering
\begin{subfigure}{.34\textwidth}
  \centering
  \includegraphics[width=.99\linewidth]{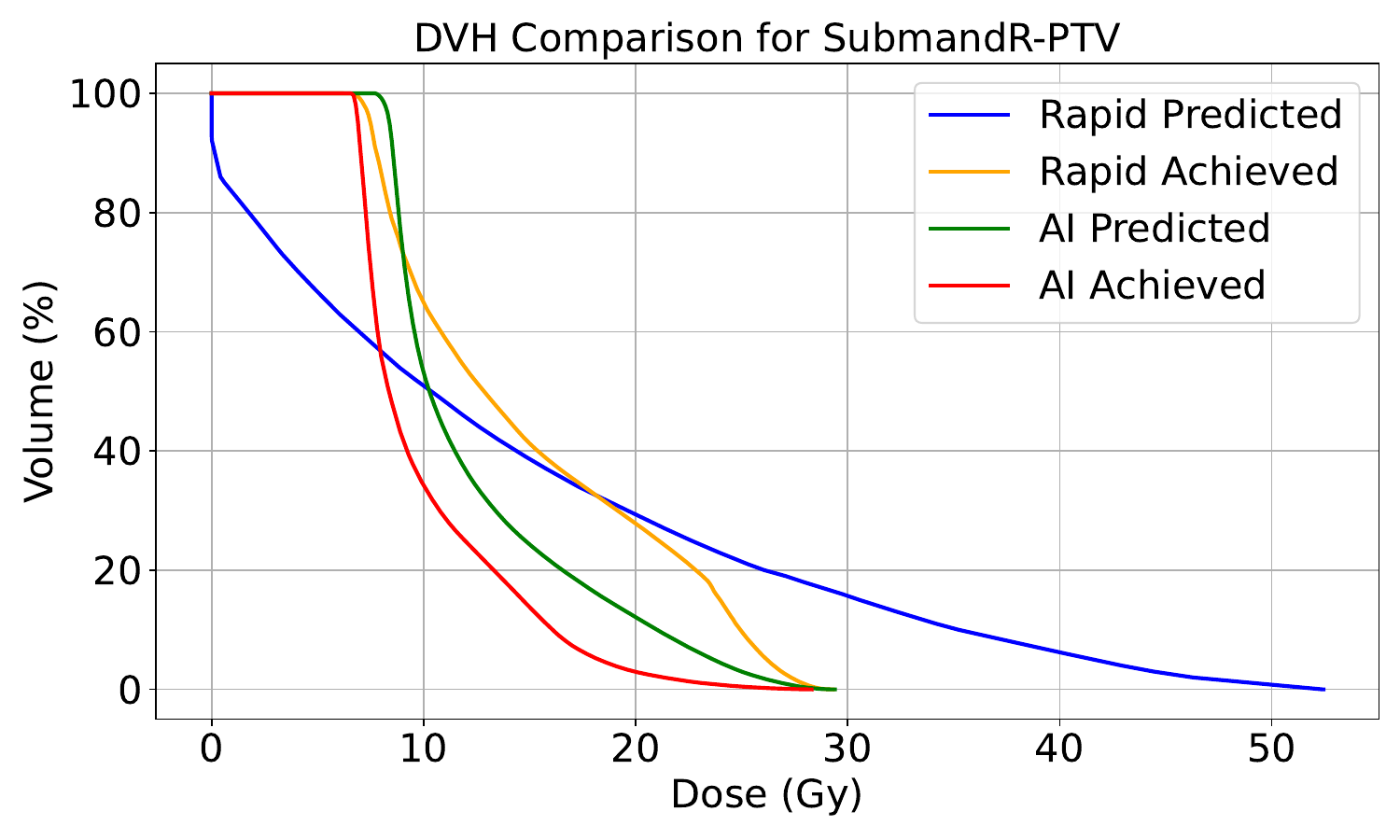}
  \label{fig:sub1}
\end{subfigure}%
\begin{subfigure}{.34\textwidth}
  \centering
  \includegraphics[width=.99\linewidth]{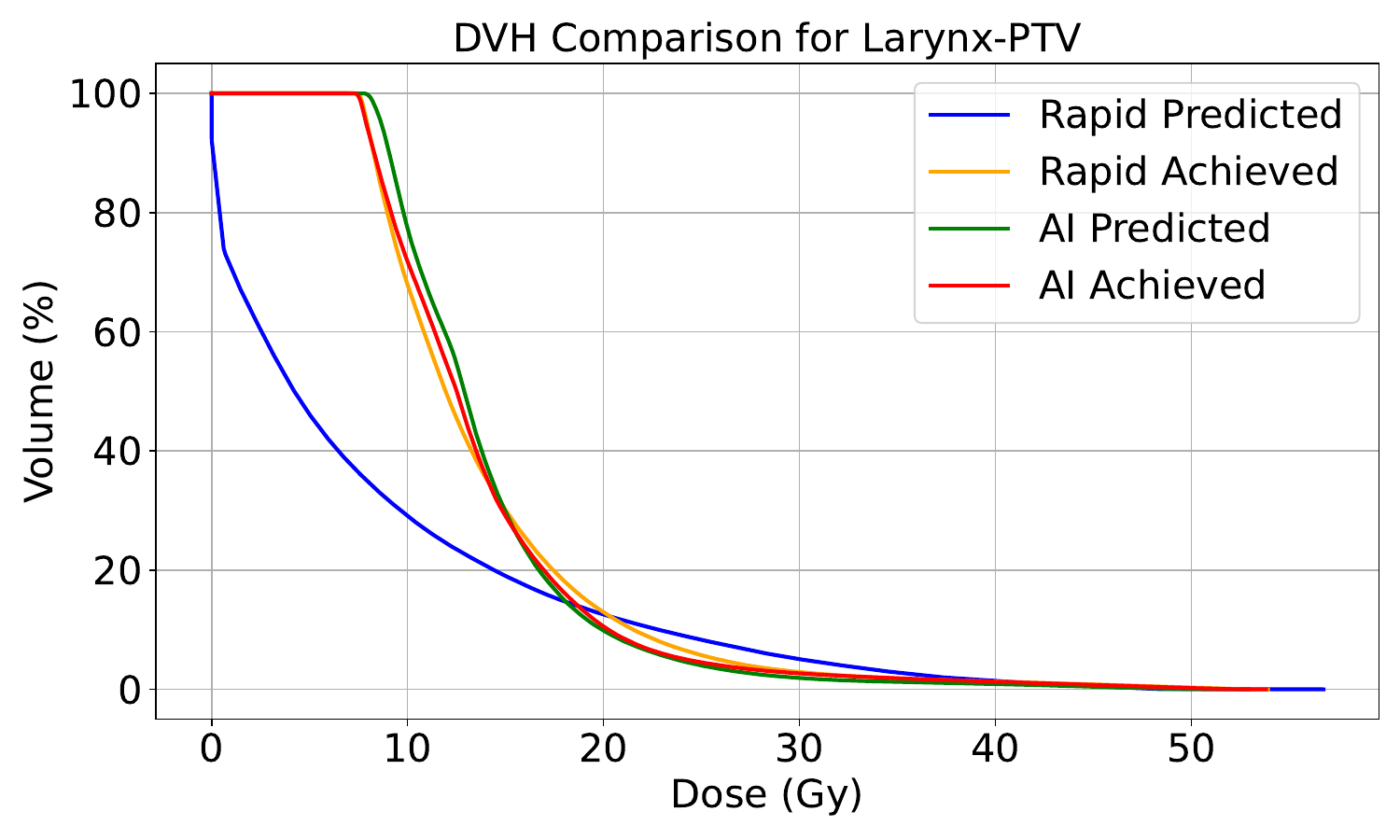}
  \label{fig:sub1}
\end{subfigure}%
\begin{subfigure}{.34\textwidth}
  \centering
  \includegraphics[width=.99\linewidth]{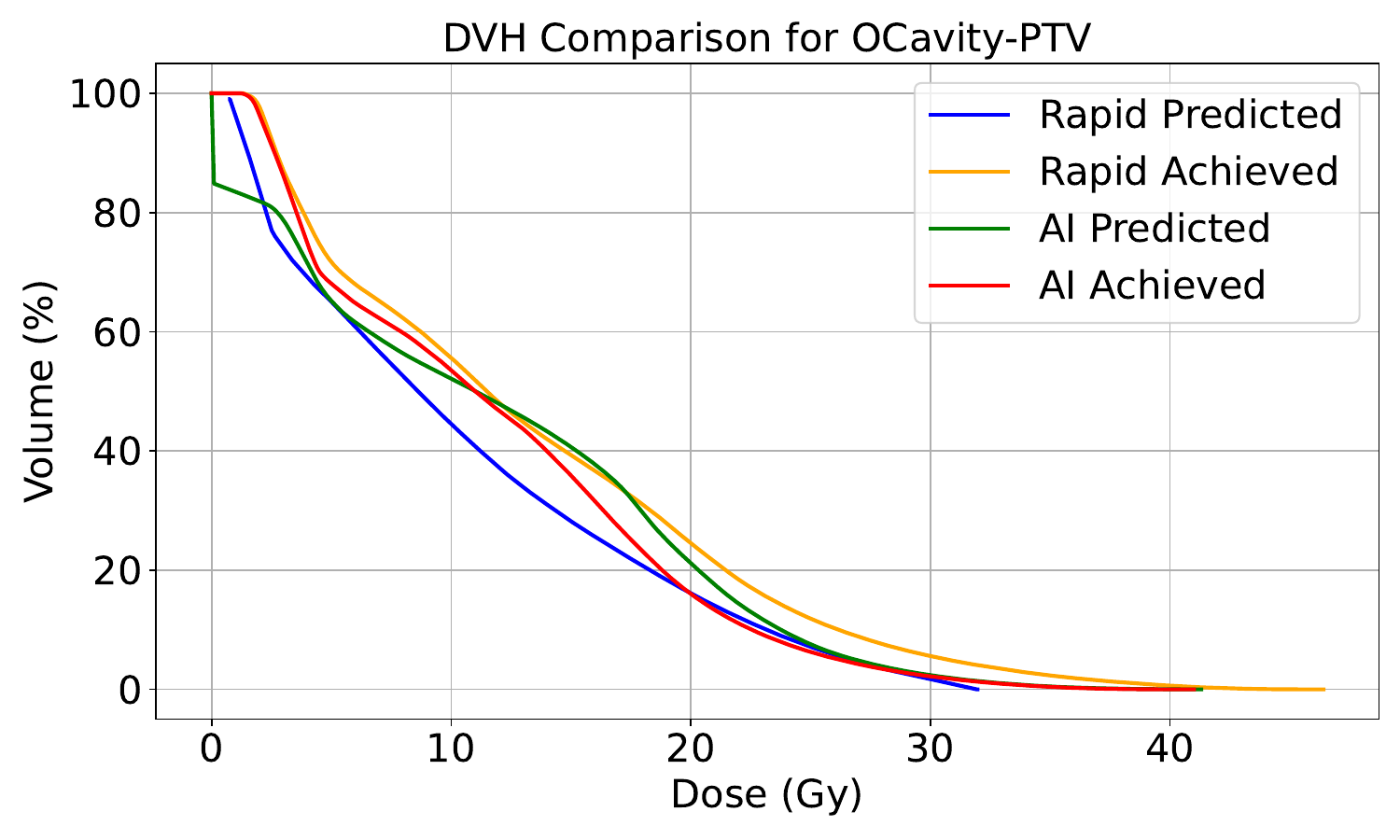}
  \label{fig:sub1}
\end{subfigure}%
\vspace{-0.1in}
\caption{Estimated vs. achieved DVHs of RapidPlan and our FDP model. RapidPlan estimations are not well aligned with achieved plans, may due to its limited generalization to unseen patients.}
\vspace{-0.1in}
\label{fig:dvhexample}
\end{figure}

Figure \ref{fig:dvhexample} shows some examples about estimated and achieved DVHs of RapidPlan and our FDP model. We notice that RapidPlan model are less robust than the proposed method, indicated by the less alignment between the blue lines and orange lines. 

Table \ref{tab:quality} shows the quality comparison of achieved plans of RapidPlan and our FDP. For OAR metrics (mean dose), FDP has achieved more larger percent on ``better" than RapidPlan for OAR sparing count (14 over 0), and 1 over 0 for PTV homogeneity \& conformity count. 
\begin{table}[]
    \centering
    \tiny
    \begin{tabular}{c|@{\hspace{6pt}}c@{\hspace{6pt}}c@{\hspace{6pt}}c@{\hspace{6pt}}c@{\hspace{6pt}}c@{\hspace{6pt}}c@{\hspace{6pt}}c@{\hspace{6pt}}c}
    \toprule
        OAR & SpinalCord 05 & Larynx-PTV & Lips & Mandible-PTV & OCavity-PTV & ParotidCon-PTV & ParotidIps-PTV & Esophagus \\
RP std & 0.98 & \textbf{2.49} & 2.69 & 3.03 & 6.34 & 1.29 & 1.61 & \textbf{0.31} \\
ours std & \textbf{0.73} & 2.76 & \textbf{1.92} & \textbf{1.05} & \textbf{3.61} & \textbf{0.56} & \textbf{0.90} & 0.32 \\
\rowcolor{gray!10}
RP mean & \textbf{0.43} & 4.24 & 2.02 & 2.22 & 6.00 & 1.66 & 2.25 & 0.23 \\
\rowcolor{gray!10}
ours mean & 0.45 & \textbf{1.55} & \textbf{1.14} & \textbf{0.83} & \textbf{3.19} & \textbf{0.64} & \textbf{0.84} & 0.23 \\
         \midrule
         OAR & SubmandL-PTV & Shoulders & SubmandR-PTV & Posterior Neck & PharConst-PTV & BrainStem 03 & Trachea & \textit{better count} \\
RP std  & 7.12 & \textbf{0.14} & 5.70 & 3.76 & 3.73 & 1.76 & 6.51 & \textit{3} \\
ours std  & \textbf{3.93} & 0.27 & \textbf{4.17} & \textbf{2.53} & \textbf{1.71} & \textbf{0.90} & \textbf{1.86} & \textit{\textbf{12}} \\
\rowcolor{gray!10}
RP mean  & 7.53 & \textbf{0.12} & 5.05 & 3.92 & 4.30 & 1.13 & 2.49 & \textit{2} \\
\rowcolor{gray!10}
ours mean  & \textbf{4.38} & 0.13 & \textbf{4.81} & \textbf{2.10} & \textbf{1.48} & \textbf{0.61} & \textbf{0.99} & \textit{\textbf{12} }\\
         \bottomrule
    \end{tabular}
    \caption{Statistics of inter-patient differences of expected and achieved DVHs ($\downarrow$). RP: RapidPlan \cite{magliari2024hn}. ours: the predicted dose from FDP and objectives extracted from AI dose for Eclipse planning.}
    \vspace{-0.1in}
    \label{tab:inter}
\end{table}

\begin{table}[]
    \centering
    \tiny
    \begin{tabular}{c|@{\hspace{6pt}}c@{\hspace{6pt}}c@{\hspace{6pt}}c@{\hspace{6pt}}c@{\hspace{6pt}}c@{\hspace{6pt}}c@{\hspace{6pt}}c@{\hspace{6pt}}c}
    \toprule
     OAR & SpinalCor 05 & Larynx-PTV & Lips & Mandible-PTV & OCavity-PTV & ParotidCon-PTV & ParotidIps-PTV & Esophagus \\
better & 47.50 & 30.00 & 47.50 & 31.25 & 64.56 & 32.50 & 48.68 & 30.23 \\
worse & 5.00 & 21.43 & 0.00 & 1.25 & 1.27 & 6.25 & 5.26 & 4.65 \\
similar & 47.50 & 48.57 & 52.50 & 67.50 & 34.18 & 61.25 & 46.05 & 65.12 \\
\midrule
OAR  & SubmandL-PTV & Shoulders & SubmandR-PTV & Posterior Neck & PharConst-PTV & BrainStem 03 & Trachea & \textbf{OAR count} \\
better  & 59.57 & 0.00 & 71.11 & 12.50 & 56.16 & 7.50 & 51.16 & \textbf{14}\\
worse  & 2.13 & 0.00 & 6.67 & 6.25 & 2.74 & 3.75 & 0.00 & \textbf{0}\\
similar & 38.30 & 100.00 & 22.22 & 81.25 & 41.10 & 88.75 & 48.84 & -\\
\midrule
PTV & HI (PTVHigh) & CI (PTVHigh) & HI (PTVMid) & CI (PTVMid) & HI (PTVLow) & CI (PTVLow) & \multicolumn{2}{c}{\textbf{PTV count}} \\
better & 0.00 & 0.00 & 0.00 & 4.00 & 1.43 & 4.29 & \multicolumn{2}{c}{\textbf{1}}  \\
worse & 0.00 & 0.00 & 0.00 & 4.00 & 0.00 & 4.29 & \multicolumn{2}{c}{\textbf{0}}  \\
similar & 100.00 & 100.00 & 100.00 & 92.00 & 98.57 & 91.43 & \multicolumn{2}{c}{-} \\
\bottomrule
    \end{tabular}
    \caption{Structure-wise plan quality comparison between RapidPlan and our FDP. Each structure is categorized as ``better", ``worse", or ``similar", indicating whether the plan generated by FDP shows improved, reduced, or comparable quality relative to RapidPlan. The ``similar" thresholds are 1 Gray for OARs; 0.015 for PTV indices (HI and CI as defined in Sec. \ref{sec:define}). Percentages (\%) are reported for each category per structure, while the \textbf{bold} column presents the column counts. }
    \label{tab:quality}
    \vspace{-0.15in}
\end{table}

\begin{figure}
    \centering
    \vspace{-0.1in}
    \includegraphics[width=\linewidth]{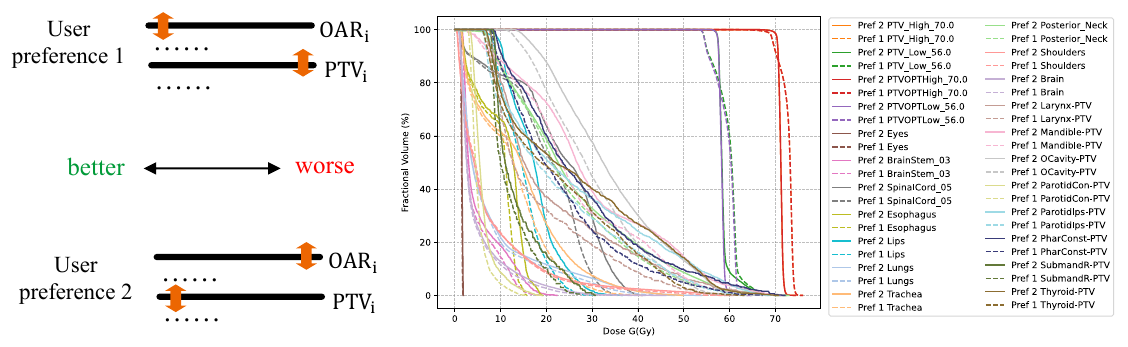}
    \caption{Demonstration of user preference with sliding bars. The left/right of sliding bars represents better/worse quality for the specific structure respectively. The Preference 1 (P1) has the preferences of OAR sparing over PTV homogenity, and Preference 2 (P2) has the opposite. The preferences have been captured in the dose prediction, as shown in the DVHs in the right panel.}
    \label{fig:slidebar}
    \vspace{-0.1in}
\end{figure}

\textbf{Sliding Bars to Adapt the Preference}. As demonstrated in Figure \ref{fig:slidebar}, we include two user preferences (P1: focusing more on OAR sparing, P2: focusing PTV over OAR).  our FDP can have dose prediction responding to the preferences. The planning system achieved plans based on different flavors have been shown in Table \ref{tab:slidebar}. More demonstration details can be found in Appendix \ref{app:interface} and \ref{sec:eclipse}. 

\begin{table}[]
    \centering
    \tiny
    \begin{tabular}{c|@{\hspace{6pt}}c@{\hspace{6pt}}c@{\hspace{6pt}}c@{\hspace{6pt}}c@{\hspace{6pt}}c@{\hspace{6pt}}c@{\hspace{6pt}}c@{\hspace{6pt}}c}
    \toprule
     OAR & SpinalCord05 & Larynx-PTV & Lips & Mandible-PTV & OCavity-PTV & ParotidCon-PTV & ParotidIps-PTV & Esophagus \\
    P1 MeanDose & 14.59 & 14.41 & 13.83 & 26.25 & 28.29 & 4.84 & 22.03 & 4.89 \\
    P2 MeanDose & 16.33 & 15.47 & 13.44 & 27.25 & 28.04 & 5.24 & 23.4 & 5.55 \\
    P1 MaxDose & 34.2 & 62.2 & 35.3 & 71.3 & 69.2 & 12.5 & 68.4 & 16.0 \\
    P2 MaxDose & 38.3 & 63.3 & 36.6 & 71.5 & 69.7 & 15.5 & 68.8 & 18.7 \\
\midrule
OAR & CochleaL & CochleaR & Shoulders & SubmandR-PTV & PosteriorNeck & PharConst-PTV & BrainStem03 & Trachea \\
P1 MeanDose & 6.28 & 1.92 & 4.13 & 10.01 & 26.04 & 19.82 & 4.64 & 9.8 \\
P2 MeanDose & 6.33 & 1.93 & 4.09 & 11.51 & 26.3 & 20.41 & 4.83 & 11.21 \\
P1 MaxDose & 9.2 & 1.9 & 61.4 & 27.5 & 73.9 & 68.8 & 21.1 & 44.2 \\
P2 MaxDose & 9.6 & 1.9 & 60.6 & 27.1 & 73.8 & 69.4 & 20.7 & 47.5 \\
\bottomrule
    \end{tabular}
    \caption{The achieved mean dose from Eclipse with different flavors based on Figure \ref{fig:slidebar}.}
    \vspace{-0.2in}
    \label{tab:slidebar}
\end{table}

\textbf{Ablation of Stage I}. In the context of user preference encoding, we find that without pre-trained Stage I, the reconstruction error is slightly larger, as shown in Table \ref{tab:test_mae} (MAE represents the mean absolute error masked by 5 Gy isodose lines). More importantly, the predicted dose is less realistic as boundary artifacts of PTVs (black arrow) and OARs (white arrow) are shown as in Figure \ref{fig:stage1}. 

\begin{minipage}{.3\linewidth}
\renewcommand{\arraystretch}{1.1}
   \small
    \centering
  \begin{tabular}{ccc}
 \toprule
  Methods  & MAE ($\downarrow$) \\
  \midrule
   w/o. S1-pretain   &  2.63  \\
   w. S1-pretain  & \textbf{2.56} \\
   \bottomrule
\end{tabular}
\captionof{table}{test set MAEs of without vs. with Stage I pretrain.}
\label{tab:test_mae}
\end{minipage}%
\hspace{0.2in}
\begin{minipage}{.65\linewidth}
\renewcommand{\arraystretch}{1.1}
 \small
    \centering
      \includegraphics[width=\linewidth]{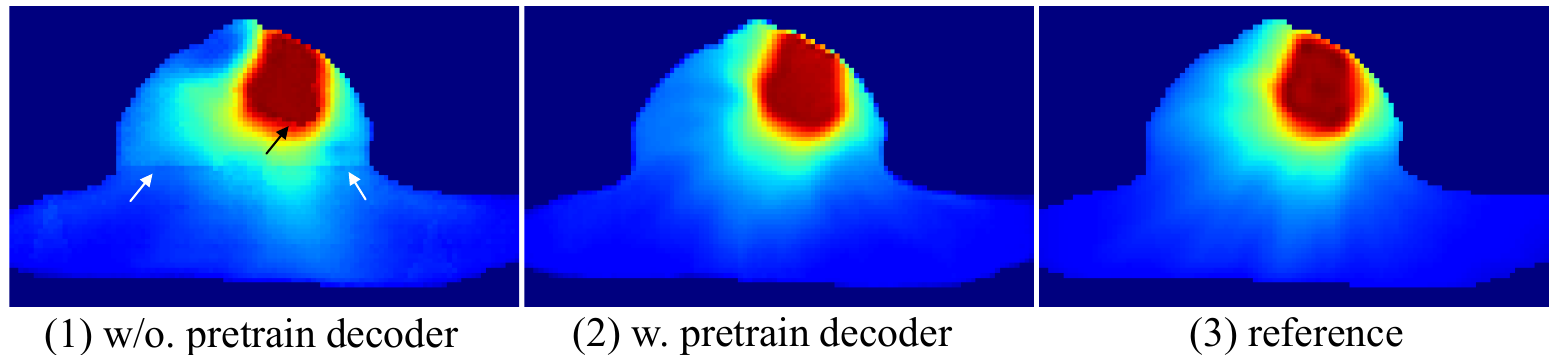}
      \label{fig:stage1}
\vspace{-0.15in}
\captionof{figure}{Example demonstrates the value of Stage I pre-training.}
\end{minipage}



\section{Discussion}

\textbf{Conclusion}. We propose a novel two-stage training framework for flexible dose prediction, representing a pioneering approach to integrate user preferences (e.g., OAR sparing, PTV homogeneity) with sliders. Unlike prior work that primarily emphasizes dose prediction accuracy, we validate the clinical utility of our AI-generated plans within a widely adopted treatment planning system. Compared to RapidPlan, our model demonstrates superior generalization in plan quality. We hope this work encourages further research and accelerates AI application for routine radiotherapy practice.

\textbf{Limitation and Future Work}. Our current evaluation focuses on head-and-neck cancer site, the most challenging region for RT planning due to its complex anatomy and stringent dose constraints. Extending our approach to other treatment sites is expected to be straightforward and is part of our future work. Second, the training paradigms of RapidPlan (e.g., conventional methods with smaller datasets) and our model (e.g., deep learning with larger-scale data) differ fundamentally, and clinical quality comparison is inherent complex and subjective. More rigorous comparisons are planned for future investigation in diverse clinical scenarios. 

\textbf{Disclaimer}. The information in this paper is based on research results that are not commercially available. Future commercial availability cannot be guaranteed. 

\bibliographystyle{abbrv}  
\bibliography{references}  



\input{appendix}
\end{document}

%% file: appendix.tex
\newpage

\section{User Interface with Sliding Bars (Demo 1)}
\label{app:interface}
\begin{figure}[h]
    \centering
    \includegraphics[width=1.05\linewidth]{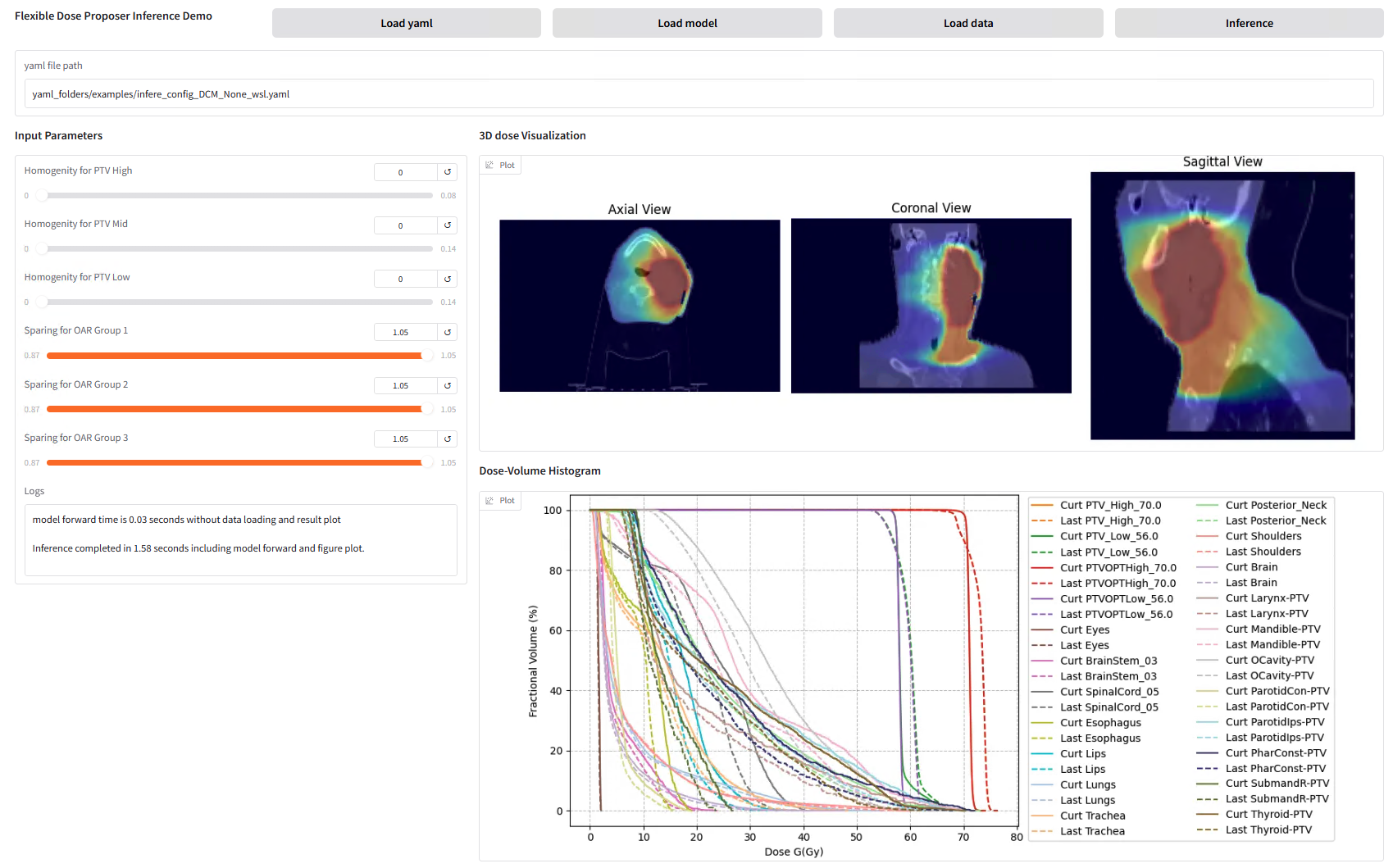}
    \caption{Screenshot of the Gradio demo interface.  A demo video recording can be found in an \href{https://huggingface.co/Jungle15/DoseProposerDemo}{HuggingFace repo}.  }
    \label{fig:demo_interface}
\end{figure}

This demo includes four steps, executed when clicking the corresponding buttons in the upper panel: 

\begin{itemize}
    \item Load yaml file. The yaml file path can be inputed by user. The yaml file includes data parameters (e.g., DICOM data paths, PTV prescribed doses, RT structures mapping) and model parameters (e.g., trained model path, model configurations, save output roots). 
    \item Load trained model. Load the trained model with PyTorch code. 
    \item Load and preprocess DICOM data. prepare the data from DICOM for deep learning input. Currently our research code is pure Python not optimized from speed, this process takes about half minutes. This is one-time running for each patient. 
    \item Inference with sliding bar parameters. The model forward only takes about 30 ms for model forward on a B200 GPU, and takes about 1.5 seconds to visualize the results. When moving the sliding and click the ``inference'' button, the results can be shown almost real-time. 
\end{itemize}

As shown in Figure \ref{fig:demo_interface}, the \textbf{upper panel} displays the buttons for the four main steps. The \textbf{left panel} presents the sliding bar interface along with the running logs. The \textbf{middle section on the right} shows the predicted dose overlaid on the CT image, while the \textbf{bottom right panel} compares the dose-volume histograms (DVHs) of the current prediction (after adjusting the sliding bars) versus the previous prediction (before sliding bars adjustment).

\newpage

\section{Deliverable Plan in Eclipse Derived from Our FDP (Demo 2)}
\label{sec:eclipse}

\begin{figure}[h]
    \centering
    \includegraphics[width=\linewidth]{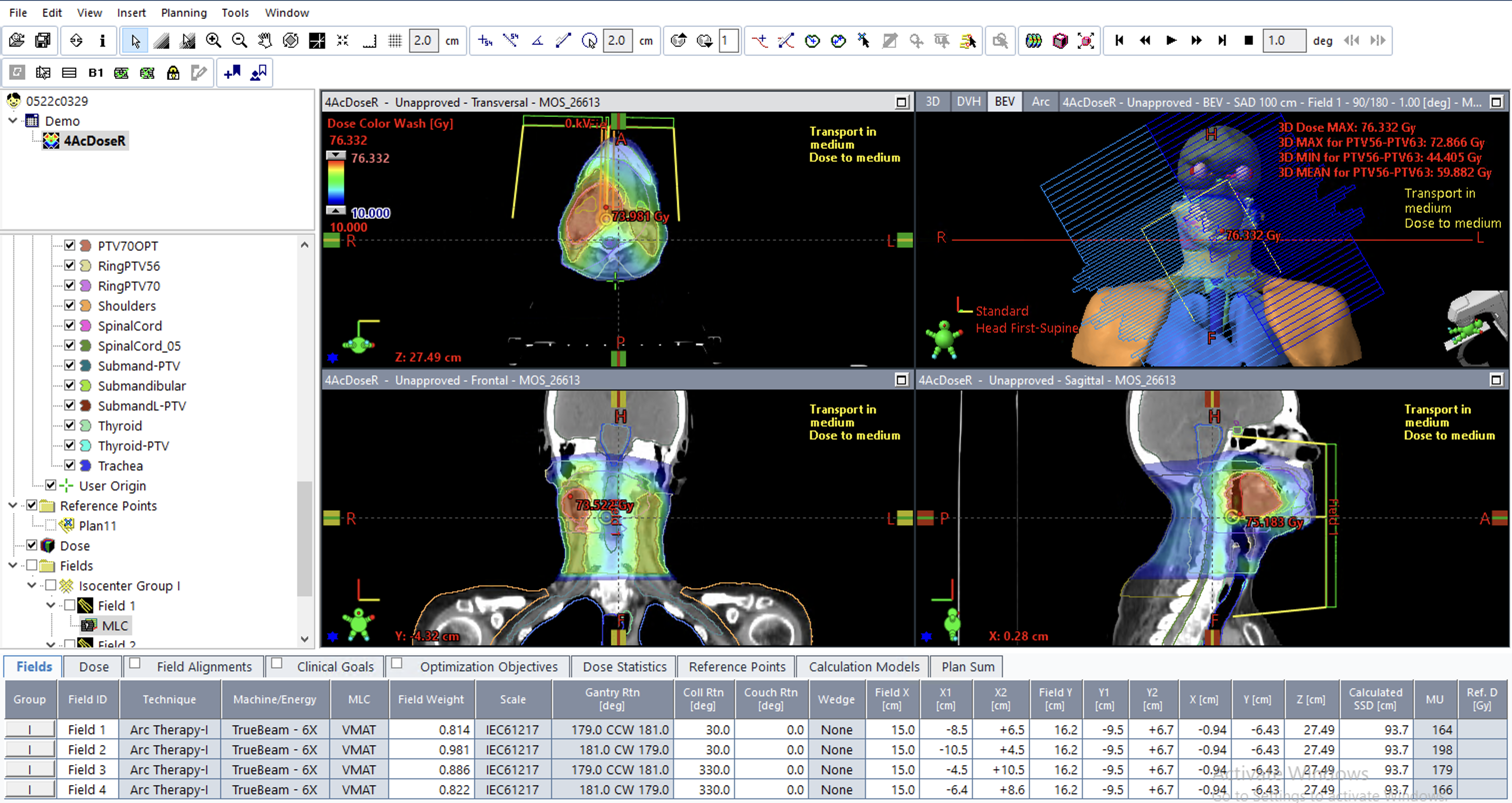}
    \caption{Screenshot of deliverable plan in Eclipse derived from our flexible dose proposer (FDP). The upper right shows the BEV of the patients with RT structures and Multi-Leaf Collimator overlaid. Other pannels show the achieved dose overlaid with CT. }
    \label{fig:eclipse}
\end{figure}

\newpage

\section{Model Structures of Our Flexible Dose Proposer}
\label{sec:structure}

Stage I model structure parameters, following the style of MONAI \footnote{\url{https://github.com/Project-MONAI/MONAI/blob/dev/monai/networks/nets/vqvae.py}}: 

\begin{table}[ht]
\centering
\caption{Stage I network structure parameters.}
\scriptsize
\begin{tabular}{@{}>{\ttfamily}l p{11cm}@{}}
\toprule
\texttt{Parameter} & \texttt{Value} \\
\midrule
\texttt{spatial\_dims} & \texttt{3} \\
\texttt{in\_channels} & \texttt{1} \\
\texttt{out\_channels} & \texttt{1} \\
\texttt{channels} & \texttt{[32, 256, 512, 512]} \\
\texttt{downsample\_parameters} & \texttt{[[2, 4, 1, 1], [2, 4, 1, 1], [1, 3, 1, 1], [1, 3, 1, 1]]} \\
\texttt{upsample\_parameters} & \texttt{[[1, 3, 1, 1, 0], [1, 3, 1, 1, 0], [2, 4, 1, 1, 0], [2, 4, 1, 1, 0]]} \\
\texttt{num\_res\_channels} & \texttt{256} \\
\texttt{num\_res\_layers} & \texttt{1} \\
\texttt{num\_embeddings} & \texttt{512} \\
\texttt{embedding\_dim} & \texttt{4} \\
\bottomrule
\end{tabular}
\label{tab:model-config}
\end{table}

Stage II image encoder structure parameters, following the style of MedNext \footnote{\url{https://github.com/MIC-DKFZ/MedNeXt/blob/main/nnunet_mednext/network_architecture/mednextv1/MedNextV1.py}}: 

\begin{table}[ht]
\centering
\caption{Stage II network structure parameters.}
\scriptsize
\begin{tabular}{@{}>{\ttfamily}l p{11cm}@{}}
\toprule
\texttt{Parameter} & \texttt{Value} \\
\midrule
\texttt{in\_channels} & \texttt{7} \\
\texttt{n\_channels} & \texttt{24} \\
\texttt{exp\_r} & \texttt{[2,3,4,4,4,4,4,3,2]} \\
\texttt{kenerl\_size} & \texttt{3} \\
\texttt{deep\_supervision} & \texttt{False} \\
\texttt{do\_res} & \texttt{True} \\
\texttt{do\_res\_up\_down} & \texttt{True} \\
\texttt{block\_counts} & \texttt{[2,3,4,4,8,8,8,4,3]} \\
\bottomrule
\end{tabular}
\label{tab:model-config}
\end{table}

\newpage
\section{Illustration and Metrics about DVH Differences}
\label{app:metrics}
\begin{figure}[h]
    \centering
    \includegraphics[width=0.9\linewidth]{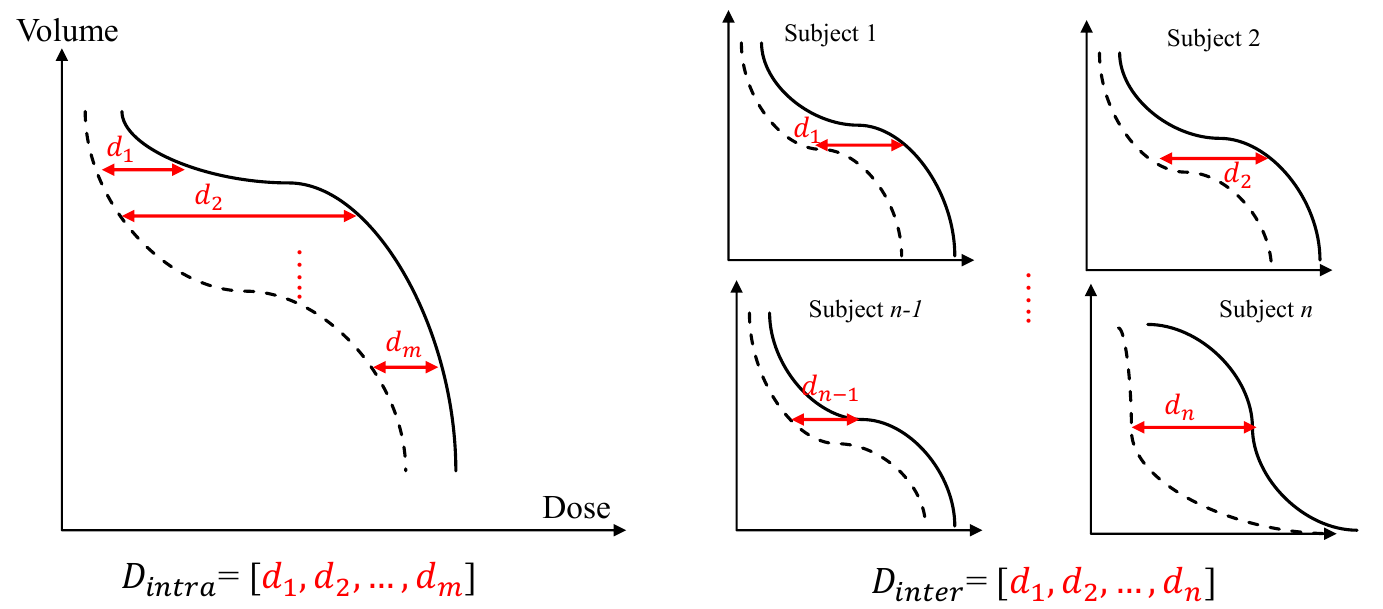}
    \caption{The cartoon illustration about intra- and inter- patient differences. }
    \label{fig:cartoon}
\end{figure}

A cartoon illustration is shown in Figure \ref{fig:cartoon} with DVHs. The solid line represents the achieved plan, and dash line represents the target (estimated) plan. Each subfigure plot the DVHs for one organ per patient. The metrics of intra- and inter- patient differences are defined below: 

\begin{align}
     \mu_\text{tra} &= \frac{1}{N} \frac{1}{L} \sum_{j=1}^{N}\sum_{i=1}^{L} D^j_i,  \sigma_\text{tra} = \frac{1}{N} \sum_{j=1}^{N}\sqrt{ \frac{1}{L} \sum_{i=1}^{L} (D^j_i - \mu^j_\text{tra})^2} \\
     \quad \mu_\text{ter} &= \frac{1}{N} \sum_{j=1}^{N} D^{j}_m,  \sigma_\text{ter} = \sqrt{ \frac{1}{N} \sum_{j=1}^{N} (D^{j}_m - \mu_\text{ter})^2}
\end{align}
where the $L$ is the length of DVH dimensions, and $N$ is the number of testing patients. 

The assumption is that DVH estimations are more accurate when the achieved DVHs are well-aligned following complex optimization using a comprehensive objective list within the Eclipse treatment planning system.

Since the mean of the differences ($D_{\text{intra}}$, $D_{\text{inter}}$) may be offset by consistent margins with some rule-based methods, so we pay less attention to the mean values in this study and shade the comparison in the tables. Constrastively, large standard deviations are indicative of poor estimation consistency and reliability.

\newpage

\section{Results when Planned with IMRT}
\label{app:imrt_plan}

The tables presented in the main text reflect results from VMAT-based planning. Below, we provide the corresponding outcomes using IMRT. The consistency across both approaches highlights the robustness of our model. 

\begin{table}[h]
    \centering
    \tiny
    \begin{tabular}{c|@{\hspace{6pt}}c@{\hspace{6pt}}c@{\hspace{6pt}}c@{\hspace{6pt}}c@{\hspace{6pt}}c@{\hspace{6pt}}c@{\hspace{6pt}}c@{\hspace{6pt}}c}
    \toprule
        OAR & SpinalCord 05 & Larynx-PTV & Lips & Mandible-PTV & OCavity-PTV & ParotidCon-PTV & ParotidIps-PTV  & Esophagus \\
        RP std & 3.92 & 2.90 & 3.15 & 2.91 & 4.21 & 3.25 & 3.09 & 3.85 \\
ours std & 1.63 & 1.71 & 1.26 & 1.49 & 2.15 & 1.06 & 1.21 & 1.63 \\
\rowcolor{gray!10}
RP mean & 5.62 & 10.75 & 5.39 & 8.08 & 8.68 & 6.33 & 7.85 & 3.89 \\
\rowcolor{gray!10}
ours mean & 2.57 & 4.33 & 1.78 & 1.99 & 3.31 & 1.92 & 1.90 & 1.69 \\
         \midrule
         OAR & SubmandL-PTV &  Shoulder &SubmandR-PTV & Posterior Neck & PharConst-PTV & Brain Stem 03 & Trachea & \textit{better count}\\
        RP std  & 3.65 & 1.59 & 2.30 & 4.64 & 3.16 & 1.49 & 3.66 & 0\\
ours std  & 1.29 & 0.57 & 1.30 & 1.25 & 1.61 & 1.02 & 1.72 &  15 \\
\rowcolor{gray!10}
RP mean  & 6.36 & 1.33 & 8.06 & 10.42 & 8.02 & 2.47 & 5.42 & 0 \\
\rowcolor{gray!10}
ours mean  & 2.96 & 0.67 & 3.53 & 2.58 & 2.48 & 1.41 & 2.83 & 15 \\
         \bottomrule
    \end{tabular}
    \caption{Statistics of intra-patient differences of expected and achieved DVHs ($ \downarrow $).}
    \label{tab:my_label}
\end{table}

\begin{table}[h]
    \centering
    \tiny
    \begin{tabular}{c|@{\hspace{6pt}}c@{\hspace{6pt}}c@{\hspace{6pt}}c@{\hspace{6pt}}c@{\hspace{6pt}}c@{\hspace{6pt}}c@{\hspace{6pt}}c@{\hspace{6pt}}c}
    \toprule
        OAR & SpinalCord 05 & Larynx-PTV & Lips & Mandible-PTV & OCavity-PTV & ParotidCon-PTV & ParotidIps-PTV  & Esophagus \\
        RP std & 2.41 & 4.44 & 2.76 & 2.72 & 5.28 & 2.20 & 2.61 & 0.57 \\
ours std & 1.92 & 2.52 & 2.25 & 0.93 & 3.62 & 1.08 & 1.05 & 0.54 \\
\rowcolor{gray!10}
RP mean & 0.88 & 8.80 & 1.98 & 3.05 & 6.07 & 2.93 & 3.66 & 0.40 \\
\rowcolor{gray!10}
ours mean & 0.79 & 4.69 & 1.41 & 0.86 & 2.91 & 0.92 & 1.10 & 0.42 \\
         \midrule
         OAR  & SubmandL-PTV &  Shoulder &SubmandR-PTV & Posterior Neck & PharConst-PTV & Brain Stem 03 & Trachea &\textit{better count} \\
        RP std  & 7.72 & 0.23 & 4.99 & 4.18 & 6.14 & 1.41 & 7.70 & 0\\
ours std  & 3.89 & 0.23 & 2.94 & 2.88 & 2.27 & 0.86 & 3.48  & 14 \\
\rowcolor{gray!10}
RP mean & 11.89 & 0.15 & 7.28 & 4.69 & 9.18 & 1.38 & 3.45 & 0\\
\rowcolor{gray!10}
ours mean  & 3.46 & 0.15 & 3.22 & 2.40 & 2.79 & 0.73 & 1.64 & 14 \\
         \bottomrule
    \end{tabular}
    \caption{Statistics of inter-patient differences of expected and achieved DVHs ($\downarrow$).}
    \label{tab:my_label}
\end{table}

\begin{table}[h]
    \centering
    \tiny
    \begin{tabular}{c|@{\hspace{6pt}}c@{\hspace{6pt}}c@{\hspace{6pt}}c@{\hspace{6pt}}c@{\hspace{6pt}}c@{\hspace{6pt}}c@{\hspace{6pt}}c@{\hspace{6pt}}c}
    \toprule
     OAR & SpinalCord 05 & Larynx-PTV & Lips & Mandible-PTV & OCavity-PTV & ParotidCon-PTV & ParotidIps-PTV & Esophagus \\
better & 36.36 & 17.65 & 37.66 & 10.39 & 50.00 & 10.39 & 13.70 & 21.95 \\
worse & 6.49 & 41.18 & 3.90 & 6.49 & 6.58 & 29.87 & 26.03 & 2.44 \\
similar & 57.14 & 41.18 & 58.44 & 83.12 & 43.42 & 59.74 & 60.27 & 75.61 \\
\midrule
OAR  & SubmandL-PTV & Shoulders & SubmandR-PTV & Posterior Neck & PharConst-PTV & BrainStem 03 & Trachea & \textbf{OAR count} \\
better  & 38.30 & 0.00 & 47.73 & 1.30 & 34.29 & 7.79 & 36.59 & \textbf{10} \\
worse  & 23.40 & 0.00 & 20.45 & 10.39 & 11.43 & 3.90 & 0.00 & \textbf{4}\\
similar & 38.30 & 100.00 & 31.82 & 88.31 & 54.29 & 88.31 & 63.41 & - \\
\midrule
PTV & HI (PTVHigh) & CI (PTVHigh) & HI (PTVMid) & CI (PTVMid) & HI (PTVLow) & CI (PTVLow) & \multicolumn{2}{c}{\textbf{PTV count}}\\
better & 5.20 & 0.00 & 13.64 & 0.00 & 2.98 & 1.49 & \multicolumn{2}{c}{\textbf{3}}\\
worse & 1.30 & 0.00 & 0.00 & 0.00 & 2.98 & 0.00  & \multicolumn{2}{c}{\textbf{0}}\\
similar & 93.51 & 100.00 & 86.36 & 100.00 & 94.03 & 98.51  & \multicolumn{2}{c}{-}\\
\bottomrule
    \end{tabular}
    \caption{Quality per structure of achieved plans of RapidPlan and our FDP. ``better", ``worse", ``similar" represent the plan derived from FDP has better, worse, or similar quality respectively. Percentages (\%) are reported for each category per structure, while the \textbf{bold} column presents the column counts. If the percent of ``better" is larger than ``worse" of one structure, the ``better" count will add 1. }
    \label{tab:placeholder}
\end{table}

\newpage 


\newpage
\section{Small Validation Sets Reflect RapidPlan Distribution More Than FDP}

\begin{table}[h]
    \centering
    \tiny
    \begin{tabular}{c|@{\hspace{6pt}}c@{\hspace{6pt}}c@{\hspace{6pt}}c@{\hspace{6pt}}c@{\hspace{6pt}}c@{\hspace{6pt}}c@{\hspace{6pt}}c@{\hspace{6pt}}c}
    \toprule
        OAR & SpinalCord05 & Larynx-PTV & Lips & Mandible-PTV & OCavity-PTV & ParotidCon-PTV & ParotidIps-PTV & Esophagus \\
RP std & 1.48 & 1.70 & 1.47 & 2.91 & 2.05 & 2.33 & 1.44 & 2.95 \\
ours std & 1.57 & 1.42 & 0.99 & 1.76 & 1.83 & 1.31 & 1.90 & 1.81 \\
\rowcolor{gray!10}
RP mean & 1.83 & 4.32 & 4.53 & 5.28 & 4.61 & 4.95 & 3.54 & 2.58 \\
\rowcolor{gray!10}
ours mean & 1.82 & 2.96 & 4.59 & 2.62 & 3.93 & 2.00 & 2.83 & 1.62 \\
         \midrule
         OAR &  SubmandL-PTV & Shoulders & SubmandR-PTV & Posterior Neck & PharConst-PTV & BrainStem03 & Trachea & better count\\
RP std  & 3.25 & 1.25 & 2.18 & 1.41 & 2.48 & 1.06 & 4.43 & 6\\
ours std & 1.58 & 0.45 & 2.65 & 1.74 & 2.66 & 1.33 & 1.91 & 9 \\
\rowcolor{gray!10}
RP mean  & 5.65 & 1.21 & 4.94 & 1.78 & 3.21 & 1.89 & 3.48 & 4\\
\rowcolor{gray!10}
ours mean  & 5.53 & 0.69 & 6.75 & 2.22 & 5.58 & 0.83 & 1.88 & 11\\
         \bottomrule
    \end{tabular}
    \caption{Statistics of intra-patient differences of expected and achieved DVHs ($ \downarrow $).}
    \label{tab:my_label}
\end{table}

\begin{table}[h]
    \centering
    \tiny
    \begin{tabular}{c|@{\hspace{6pt}}c@{\hspace{6pt}}c@{\hspace{6pt}}c@{\hspace{6pt}}c@{\hspace{6pt}}c@{\hspace{6pt}}c@{\hspace{6pt}}c@{\hspace{6pt}}c}
    \toprule
        OAR & SpinalCord05 & Larynx-PTV & Lips & Mandible-PTV & OCavity-PTV & ParotidCon-PTV & ParotidIps-PTV & Esophagus \\
RP std & 0.12 & 0.29 & 1.37 & 1.16 & 4.49 & 1.11 & 0.66 & 0.39 \\
ours std & 0.12 & 0.41 & 0.67 & 0.85 & 1.87 & 0.63 & 0.68 & 0.39 \\
\rowcolor{gray!10}
RP mean & 0.38 & 4.23 & 4.93 & 2.80 & 9.90 & 2.61 & 2.59 & 0.38 \\
\rowcolor{gray!10}
ours mean & 0.37 & 0.72 & 4.09 & 1.11 & 2.69 & 0.98 & 1.00 & 0.36 \\
         \midrule
         OAR  & SubmandL-PTV & Shoulders & SubmandR-PTV & PosteriorNeck & PharConst-PTV & BrainStem03 & Trachea & better count\\        
RP std & 0.14 & 0.08 & 3.14 & 0.13 & 2.14 & 0.00 & 0.53 & 5\\
ours std & 0.57 & 0.10 & 3.94 & 0.10 & 0.97 & 0.00 & 0.36 & 7\\
\rowcolor{gray!10}
RP mean  & 8.29 & 0.15 & 4.20 & 0.21 & 5.33 & 1.05 & 1.18 & 3 \\
\rowcolor{gray!10}
ours mean  & 3.44 & 0.15 & 6.06 & 0.24 & 1.61 & 1.24 & 0.70 & 11 \\
         \bottomrule
    \end{tabular}
    \caption{Statistics of inter-patient differences of expected and achieved DVHs ($\downarrow$).}
    \label{tab:my_label}
\end{table}

\begin{table}[h]
    \centering
    \tiny
    \begin{tabular}{c|@{\hspace{6pt}}c@{\hspace{6pt}}c@{\hspace{6pt}}c@{\hspace{6pt}}c@{\hspace{6pt}}c@{\hspace{6pt}}c@{\hspace{6pt}}c@{\hspace{6pt}}c}
    \toprule
     OAR & SpinalCord 05 & Larynx-PTV & Lips & Mandible-PTV & OCavity-PTV & ParotidCon-PTV & ParotidIps-PTV & Esophagus \\
better & 25.00 & 0.00 & 50.00 & 75.00 & 75.00 & 25.00 & 50.00 & 50.00 \\
worse & 0.00 & 50.00 & 0.00 & 25.00 & 0.00 & 0.00 & 0.00 & 0.00 \\
similar & 75.00 & 50.00 & 50.00 & 0.00 & 25.00 & 75.00 & 50.00 & 50.00 \\
\midrule
OAR  & SubmandL-PTV & Shoulders & SubmandR-PTV & Posterior Neck & PharConst-PTV & BrainStem03 & Trachea & \textbf{OAR count} \\        
better  & 50.00 & 0.00 & 66.67 & 0.00 & 100.00 & 0.00 & 75.00 & \textbf{10} \\
worse  & 50.00 & 0.00 & 0.00 & 25.00 & 0.00 & 0.00 & 0.00 & \textbf{2} \\
similar & 0.00 & 100.00 & 33.33 & 75.00 & 0.00 & 100.00 & 25.00 & -\\
\midrule
PTV & high homo & high conf & mid homo & mid conf & low homo & low conf & \multicolumn{2}{c}{\textbf{PTV count}} \\
better & 0.00 & 0.00 & 0.00 & 0.00 & 0.00 & 0.00 & \multicolumn{2}{c}{\textbf{0}}\\
worse & 0.00 & 0.00 & 0.00 & 0.00 & 0.00 & 0.00 & \multicolumn{2}{c}{\textbf{0}}\\
similar & 100.0 & 100.0 & 100.0 & 100.0 & 100.0 & 100.0 & \multicolumn{2}{c}{-}\\
\bottomrule
    \end{tabular}
    \caption{Quality per structure of achieved plans of RapidPlan and our FDP. ``better", ``worse", ``similar" represent the plan derived from FDP has better, worse, or similar quality respectively. Percentages (\%) are reported for each category per structure, while the \textbf{bold} column presents the column counts. If the percent of ``better" is larger than ``worse" of one structure, the ``better" count will add 1.}
    \label{tab:placeholder}
\end{table}

\newpage

\section{Translate 3D Dose Prediction to Optimization Objectives}

\label{sec:dose2obj}

\begin{figure}[h]
    \centering
    \includegraphics[width=0.95\textwidth]{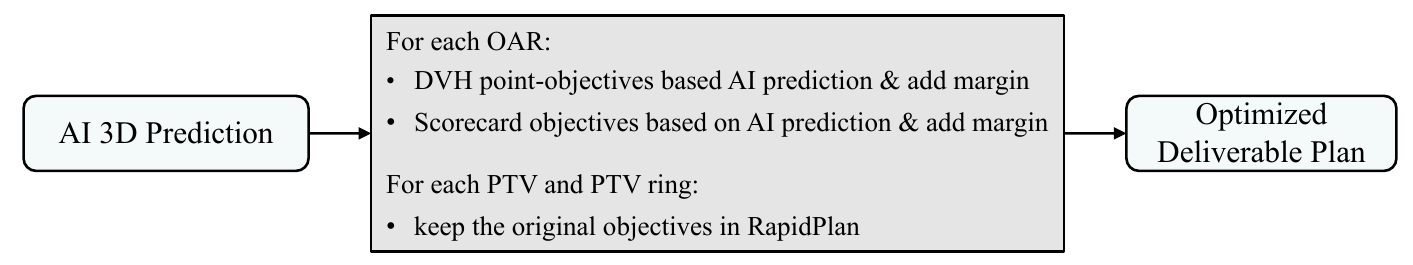}
    \caption{Illustration of translating 3D dose prediction to optimization objectives in Eclipse TPS.}
    \label{fig:dose2obj}
\end{figure}

Currently, Eclipse TPS does not support the optimization with 3D dose distribution as the direct objective. Therefore, we use a method to convert the 3D dose prediction into a set of dose-volume and mean dose objectives that can be utilized within Eclipse TPS for plan optimization. As shown in Figure \ref{fig:dose2obj}, for each organ at risk (OAR) we sample the predicted dose-volume histogram (DVH) at specific volume percentages to create corresponding dose-volume objectives after adding some rule-based margins. Additionally, based on metrics from a scorecard \cite{varian2024han}, we further adding mean dose and dose-volume objectives to better align with clinical goals. 

For PTVs and PTV rings, we follow the settings of the RapidPlan \cite{varian2024rapidmodels}, where the objectives are not from a model estimation but from dose prescriptions.